\documentclass[lettersize,journal,twoside]{IEEEtran}
\usepackage{amsmath,amsfonts}
\usepackage{algorithm}
\usepackage{array}
\usepackage{textcomp}
\usepackage{stfloats}
\usepackage{url}
\usepackage{verbatim}
\usepackage{graphicx}
\usepackage{cite}
\usepackage{amssymb}
\usepackage{algorithmic}

\usepackage[font=footnotesize,labelfont=sc,labelsep=period]{caption}
\usepackage{balance}
\usepackage{multirow,tabularx}
\usepackage{CJKutf8}
\usepackage[export]{adjustbox}
\usepackage{calc}
\usepackage[utf8]{inputenc}
\usepackage[colorinlistoftodos]{todonotes}
\usepackage{color}
\usepackage{booktabs}
\usepackage{pifont}
\usepackage{afterpage}
\usepackage{colortbl}
\usepackage{makecell}
\usepackage[table,xcdraw]{xcolor}
\usepackage{hyperref}
\usepackage{soul}
\usepackage[caption=false,font=footnotesize,labelfont=rm,textfont=rm]{subfig}
\usepackage{xspace}

\hypersetup{
  colorlinks=true,
  linkcolor=blue,
  urlcolor=cyan,
}

\def\BibTeX{{\rm B\kern-.05em{\sc i\kern-.025em b}\kern-.08em
    T\kern-.1667em\lower.7ex\hbox{E}\kern-.125emX}}

\newcommand{\xmark}{\ding{55}}

\definecolor{Gray}{gray}{0.85}
\definecolor{Tgray}{gray}{0.85}
\sethlcolor{Tgray}

\def\secref#1{Section~\ref{#1}}
\def\figref#1{Fig.~\ref{#1}}

\def\tabref#1{Table~\ref{#1}}
\def\eqref#1{(\ref{#1})}

\def\vstabcap{\vspace{-0.16cm}}
\def\vstabfoot{\vspace{0.07cm}}

\def\vsequ{\vspace{-0.15cm}}

\newcommand{\rom}[1]{\uppercase\expandafter{\romannumeral #1\relax}}
\newcommand{\romsmall}[1]{\lowercase\expandafter{\romannumeral #1\relax}}
\makeatletter
\DeclareRobustCommand\onedot{\futurelet\@let@token\@onedot}
\def\@onedot{\ifx\@let@token.\else.\null\fi\xspace}

\makeatother

\newcolumntype{L}[1]{>{\raggedright\let\newline\\\arraybackslash\hspace{0pt}}m{#1}}
\newcolumntype{C}[1]{>{\centering\let\newline\\\arraybackslash\hspace{0pt}}m{#1}}
\newcolumntype{R}[1]{>{\raggedleft\let\newline\\\arraybackslash\hspace{0pt}}m{#1}}
\newcolumntype{N}{>{\raggedleft\arraybackslash}p{0.80cm}}

\begin{document}

\bstctlcite{IEEEexample:BSTcontrol}
\title{ASPIRE-VINS: Adaptive Spline-based Visual-inertial Navigation System With Robust 3D Measurement Residuals
}
\author{
% Anonymous authors for double-blind review
Kwangyik Jung, Eungchang Mason Lee, \textit{Member, IEEE}, Taekjun Oh, \\and Hyun Myung, \textit{Senior Member, IEEE}
}
\markboth{IEEE Robotics and Automation Letters. Preprint Version. Accepted June, 2026}%
{Jung \MakeLowercase{\textit{et al.}}: ASPIRE-VINS: Adaptive Spline-based Visual-inertial Navigation System With Robust 3D Measurement Residuals}
% {How to Use the IEEEtran \LaTeX \ Templates}

\maketitle
\begin{abstract}
Visual-inertial navigation systems estimate six-degree-of-freedom motion by fusing visual and inertial data.
Modern discrete-time methods with IMU preintegration provide strong accuracy and efficiency, but keyframe-based representations can be less flexible when residuals must be evaluated at arbitrary timestamps or when motion-dependent temporal resolution is needed.
Continuous-time splines address this issue by representing the trajectory as a smooth temporal function, but uniformly spaced knots can under-represent rapid dynamics or over-parameterize static intervals.
This letter proposes ASPIRE-VINS, a continuous-time VINS framework that combines adaptive knot placement (AKP), multi-resolution splines (MRS), and 3D measurement-space residuals (3D-MSR).
AKP allocates knots according to local motion variation, MRS adds bounded local refinement in tangent space, and 3D-MSR provides bearing consistency by aligning transformed features with calibrated observation rays in 3D measurement space.
Experiments show that ASPIRE-VINS achieves competitive or lower trajectory errors than the compared baselines, demonstrating the effectiveness of motion-adaptive continuous-time trajectory modeling under diverse motion and sensing conditions.
\end{abstract}


%전체 흐름을 기존 한계 → continuous-time의 등장과 장점 → 여전히 남아 있는 한계(균일 knot, 2D residual) → ASPIRE-VINS의 해결책과 기여
\section{INTRODUCTION}
\IEEEPARstart{V}{isual}-inertial navigation systems (VINS) estimate six-degree-of-freedom trajectories by fusing camera and inertial measurement unit (IMU) data.
Discrete-time frameworks, such as MSCKF~\cite{sun2018robust}, VINS-Mono~\cite{qin2018vins}, PL-VINS~\cite{fu2020plvins}, and recent square-root filters such as $\sqrt{\mathrm {VINS}}$~\cite{peng2025sqrt}, remain strong real-time baselines.
They achieve practical accuracy and efficiency, but their state updates are tied to discrete timestamps, which can be less flexible when residuals are evaluated at native sensor times or when trajectory resolution should vary with motion intensity.

Continuous-time representations address this issue by modeling motion as a smooth function of time~\cite{furgale2012continuous, bourmaud2015continuous, mueggler2018continuous, lang2022ctrl, hug2022continuous, dellenbach2022ct}.
They support timestamp-continuous residual evaluation for high-rate IMU and low-rate camera data.
However, many continuous-time systems use uniformly spaced knots, which can underfit rapid motion or add redundant parameters in low-dynamic intervals.

Many VINS frameworks also rely on 2D image-plane reprojection errors for visual constraints.
Although effective, these residuals can be sensitive to pixel noise, feature localization errors, and visual degradation during rapid motion~\cite{qin2018vins, he2018plvio, fu2020plvins, du2024sp, tran2025robust, song2024dynavins++}.
This motivates residual formulations that operate in 3D measurement space while still depending on calibrated bearing vectors.

In this paper, we present \textbf{ASPIRE-VINS}, an 
\textbf{A}daptive \textbf{S}pline-based 
\textbf{P}latform for \textbf{I}nertial–visual 
\textbf{R}obust \textbf{E}stimation, which integrates three components. 

\begin{enumerate}
	\item \textbf{Adaptive knot placement (AKP)} dynamically allocates spline knots according to local motion variation, providing higher temporal resolution in fast or nonlinear segments while avoiding redundancy in static intervals. AKP therefore determines where the trajectory should be represented with finer detail.
	
	\item \textbf{Multi-resolution splines (MRS)} model the trajectory across multiple temporal scales, enabling coarse levels to maintain global smoothness while fine levels refine localized dynamics identified by AKP. This hierarchy yields a compact yet expressive continuous-time representation that adapts to varying motion complexity. 
	
	\item \textbf{3D measurement-space residuals (3D-MSR)} enforce feature alignment directly in Euclidean 3D space by constraining predicted points to lie along the observation rays of their corresponding measurements. Evaluated at the timestamps provided by the continuous-time spline, these residuals reduce direct reliance on image-plane reprojection.
    % while still using calibrated bearing vectors.
\end{enumerate}

Together, these components form a unified continuous-time optimization framework in which spline control points across all resolution levels serve as the optimization variables, and 3D-MSR provides temporally accurate geometric constraints along the continuous trajectory. The resulting formulation improves motion adaptability and visual residual consistency under the evaluated conditions, including nonlinear motion and visual degradation.

% 문단 간 연결 강화: Discrete-time → Continuous-time 초기 연구 → OKVIS/CTRL 한계 → CT-UIO 보완점과 한계 → 2D residual의 취약성 → 3D residual 기반 연구 → ASPIRE-VINS 제안 의 흐름으로 구성.
\section{Related Works}
Discrete-time VINS approaches, such as MSCKF~\cite{sun2018robust}, VINS-Mono~\cite{qin2018vins}, and SPVIO~\cite{du2024sp}, optimize or filter over keyframe-based or frame-wise poses.
They provide real-time performance and strong accuracy, but their updates are tied to selected timestamps or image frames.
% Feature-enhanced systems, including PL-VIO~\cite{he2018plvio} and PL-VINS~\cite{fu2020plvins}, add line features while retaining image-domain residuals.
Recent filtering methods have further improved efficiency and numerical stability through equivariant filtering in EqVIO~\cite{van2023eqvio}, Schur-complement marginalization in SchurVINS~\cite{fan2024schurvins}, and square-root covariance propagation in $\sqrt{\mathrm {VINS}}$~\cite{peng2025sqrt}.
These discrete-time methods provide strong practical baselines, whereas continuous-time trajectories offer a complementary representation for evaluating inter-frame states and adapting temporal resolution to motion variation.

Continuous-time models improve temporal modeling beyond discrete-state formulations.
Furgale \textit{et al.}~\cite{furgale2012continuous} and Mueggler \textit{et al.}~\cite{mueggler2018continuous} used cubic B-splines in \(SE(3)\) to interpolate sensor trajectories and integrate measurements at different rates.
OKVIS-CT~\cite{hug2022continuous} and Ctrl-VIO~\cite{lang2022ctrl} further embedded Lie group splines into full state-estimation systems.
However, their uniformly spaced knots can limit representation during highly dynamic intervals or add unnecessary parameters during static intervals~\cite{mo2021continuous,talbot2025continuous}.

CT-UIO~\cite{sun2025ct} addressed uniform-knot inefficiency using non-uniform B-splines with adaptive knot placement from local motion dynamics.
It reduced redundancy in motion-variable environments, but it was developed for UWB-inertial odometry and does not include visual feature-level residuals.
Its Euclidean-space representation and decoupled fitting process are also not directly tailored to tightly coupled visual-inertial optimization.

Many VINS frameworks rely on 2D image-plane reprojection errors, which can degrade under pixel noise, feature localization errors, and visual degradation~\cite{qin2018vins,he2018plvio,fu2020plvins}.
PL-VIO~\cite{he2018plvio} and PL-VINS~\cite{fu2020plvins} enrich image-domain constraints with line features, and UV-SLAM~\cite{lim2022uv} introduces uncertainty-aware visual constraints.
However, these methods still depend on projective residuals.

Recent studies have extended 3D residual modeling to visual-inertial systems.
DiT-SLAM~\cite{dit2022slam} uses implicit 3D depth in graph optimization but increases real-time burden.
D-VINS~\cite{dvins2023adaptive} uses semantic cues and dynamic-object filtering, while GaussianFlow SLAM~\cite{seo2026gaussianflow} and AIM-SLAM~\cite{jeon2026aim} rely on dense or learned front-end priors and remain sensitive to front-end estimation quality.
A, B, VINS~\cite{abvins2024simple} uses Euclidean residuals for efficient \(SE(3)\) optimization, but its minimal design can be less robust under nonlinear motion or visual degradation.

ASPIRE-VINS integrates adaptive spline resolution and 3D measurement-space residuals in one continuous-time optimization.
AKP and MRS reduce the limitations of uniform discretization, while 3D-MSR enforces feature alignment in Euclidean 3D space instead of relying solely on pixel-level reprojection.
This formulation positions ASPIRE-VINS as a motion-adaptive continuous-time backend that combines non-uniform trajectory representation with calibrated bearing-vector constraints.

\section{Adaptive Spline-\MakeLowercase{b}ased Visual-inertial Navigation System with Robust 3D Measurement Residuals}
\begin{figure}[!t]
 \vspace{-0.14cm}
    	\centering
	\includegraphics[width=0.23\textwidth]{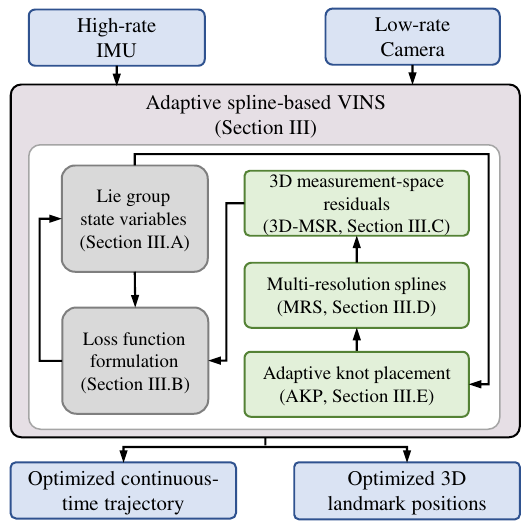}
    \caption{Compact pipeline of ASPIRE-VINS. AKP determines the knot distribution, MRS evaluates the continuous-time trajectory on $SE(3)$, and IMU and 3D-MSR residuals are jointly optimized.}
    \label{fig:method_pipeline}
    \vspace{-0.9cm}
\end{figure}

VINS generally follows a prediction-update pipeline: high-rate IMU measurements propagate the state, and lower-rate camera frames provide drift-correcting observations.
ASPIRE-VINS preserves this principle in continuous time by estimating the trajectory, velocity, inertial biases, and 3D landmarks with the Lie group formulation of MSCKF-DVIO~\cite{jung2024msckf} and a continuous-time B-spline representation.
This representation provides smooth interpolation, analytic inertial derivatives, and visual constraints in 3D measurement space, allowing inertial and visual cues to be fused in one optimization.
\figref{fig:method_pipeline} summarizes the proposed backend.

\subsection{State Variables}
\label{sec:state_variables}
The system state \(\mathbf{X}\) contains the continuous-time pose trajectory \(\mathbf{T}_{w}^{s}(t)\), which represents the pose of the sensor frame $s$ with respect to the world frame $w$ at time $t$. The trajectory is modeled by spline coefficients in local \(SE(3)\) coordinates and is mapped back to \(SE(3)\) by retraction. The state also includes the linear velocity $\mathbf{v}(t)\in\mathbb{R}^3$, accelerometer and gyroscope biases $\mathbf{b}_a(t),\mathbf{b}_g(t)\in\mathbb{R}^3$, and observed 3D landmark positions $\mathcal{P}$ as follows:
\vsequ \begin{equation}
	\begin{split}
		&\mathbf{X} = \{ \mathbf{T}_{w}^{s}(t), \mathbf{v}(t), \mathbf{b}_a(t), \mathbf{b}_g(t), \mathcal{P} \},
		\\
		& 
		\mathcal{P} = \{\mathbf{P}_1, \dots, \mathbf{P}_N\}\in\mathbb{R}^{3N}.
	\end{split}
	\label{eqa:StateVariables}
\end{equation}
\noindent The full state space is embedded in \(SE(3) \times \mathbb{R}^{9+3N}\).
The angular velocity $\boldsymbol{\omega}(t)$ is not treated as an independent state.
Instead, it is obtained from the first time derivative of the rotation spline embedded in $\mathbf{T}_{w}^{s}(t)\in SE(3)$, following continuous-time visual-inertial formulations.
This avoids introducing angular velocity as a redundant state variable.

For optimization, we employ the Lie group retraction used in MSCKF-DVIO to ensure consistency and efficient Jacobian computation. The spline interpolation and optimizer increments are represented in local tangent coordinates rather than by a direct Euclidean average of transformation matrices. State updates are performed in the tangent space as follows:
\vsequ \begin{equation}
	\begin{split}
		\mathbf{x} \oplus \delta\boldsymbol{\xi} = \exp(\delta\boldsymbol{\xi}) \cdot \mathbf{x},
	\end{split}
\end{equation}
\noindent where \(\delta\boldsymbol{\xi} \in \mathfrak{se}(3) \times \mathbb{R}^9\) denotes a minimal perturbation, and \(\mathbf{x}\) is the state in the tangent space, excluding landmark positions.

\subsection{Loss Function Formulation}
\label{sec:cost_function_formulation}
The state variables are optimized by minimizing a loss function that fuses inertial and visual data while regularizing physically plausible motion.
The overall loss function \(\mathcal{J}(\mathbf{X})\) is formulated as follows:
\vsequ \begin{equation}
    \begin{split}
\mathcal{J}(\mathbf{X}) = & \sum_{i} \Big\{ \| \boldsymbol{\omega}_i - \hat{\boldsymbol{\omega}}_i \|^2_{\Sigma_{\omega}} + \| \mathbf{a}_i - \hat{\mathbf{a}}_i \|^2_{\Sigma_a} \Big\} \\
& + \sum_{j} \left\|
\mathbf{f}_P\!\left(
\mathbf{P}_{m_j}, \mathbf{P}_{r};
\mathbf{T}_r^{m_j}, \mathbf{T}_w^s(t_j)
\right)
\right\|^2_{\Sigma_P} \\
& + \lambda \int \left(
\| \dot{\boldsymbol{\omega}}(t) \|^2_{\Sigma_{\dot{\omega}}}
+ \| \ddot{\mathbf{p}}(t) \|^2_{\Sigma_{\ddot{p}}}
\right) dt .
    \end{split}
    \label{eqa:FULLCOST}
\end{equation}
\noindent Here, \(\boldsymbol{\omega}_i\) and \(\mathbf{a}_i\) denote the bias-compensated gyroscope and accelerometer measurements at timestamp \(t_i\), while \(\hat{\boldsymbol{\omega}}_i\) and \(\hat{\mathbf{a}}_i\) are the corresponding spline-based predictions.
The prediction \(\hat{\boldsymbol{\omega}}_i\) is obtained from the right-trivialized derivative of the rotation spline, and \(\hat{\mathbf{a}}_i\) is computed from the translational component of \(\mathbf{T}_w^s(t)\) expressed in the sensor frame.
The covariance matrices \(\Sigma_{\omega}\) and \(\Sigma_a\) weight the inertial residuals.

Each visual measurement contributes to the 3D measurement-space residual \(\mathbf{f}_P(\cdot)\).
For the \(j\)-th observation, \(\mathbf{P}_{m_j}\) denotes the back-projected measurement ray direction obtained from the calibrated camera model, and \(\mathbf{P}_{r}\in\mathbb{R}^3\) denotes the associated reference 3D feature point.
The residual is evaluated using the relative transformation \(\mathbf{T}_r^{m_j} \in SE(3)\) and the interpolated pose \(\mathbf{T}_w^s(t_j) \in SE(3)\).
The details of \(\mathbf{f}_P(\cdot)\) and \(\mathbf{T}_w^s(t_j)\) are provided in \secref{sec:3d_msr} and \secref{sec:mrs}, respectively, and \(\Sigma_P\) is the visual-measurement covariance.

The final term is a tangent-space smoothness prior.
Here, \(\dot{\boldsymbol{\omega}}(t)\) is the time derivative of the body angular velocity, \(\mathbf{p}(t)\) is the translational component of \(\mathbf{T}_w^s(t)\), and \(\ddot{\mathbf{p}}(t)\) is its second derivative.
The covariance matrices \(\Sigma_{\dot{\omega}}\) and \(\Sigma_{\ddot{p}}\) weight angular-velocity variation and translational-acceleration regularization, respectively.
This prior suppresses implausible motion without directly differentiating the homogeneous transformation matrix.
The hyperparameter \(\lambda\) balances measurement fidelity and trajectory regularization, and each residual is weighted by the inverse of its covariance.
The resulting loss extends the Lie group retraction of MSCKF-DVIO~\cite{jung2024msckf} to non-uniform multi-resolution splines and is solved using Ceres~\cite{agarwal2012ceres}.
% Throughout this section, the index \(i\) refers to IMU measurement timestamps, while \(j\) indexes visual measurements and their corresponding timestamps.
Throughout this section, \(i\) indexes IMU samples, whereas \(j\) indexes visual measurements.

\subsection{Visual Residuals: 3D Measurement-Space Residuals (3D-MSR)} 
\label{sec:3d_msr}
To incorporate visual information into the continuous-time framework, we formulate the visual residual in 3D measurement space.
Unlike conventional VINS methods~\cite{qin2018vins, he2018plvio, fu2020plvins} that rely on 2D image-plane reprojection errors, 3D-MSR evaluates the consistency between a transformed reference feature and its back-projected measurement ray.
This formulation minimizes the ray-orthogonal component defined by calibrated bearing vectors and reduces direct reliance on pixel-plane reprojection under visually degraded conditions~\cite{dit2022slam, abvins2024simple}.
It assumes calibrated observations in the corresponding sensor frame and does not compensate for calibration errors or feature mismatches.

\begin{figure}[!t]
 \vspace{-0.14cm}
	\centering
	\includegraphics[width=0.29\textwidth]{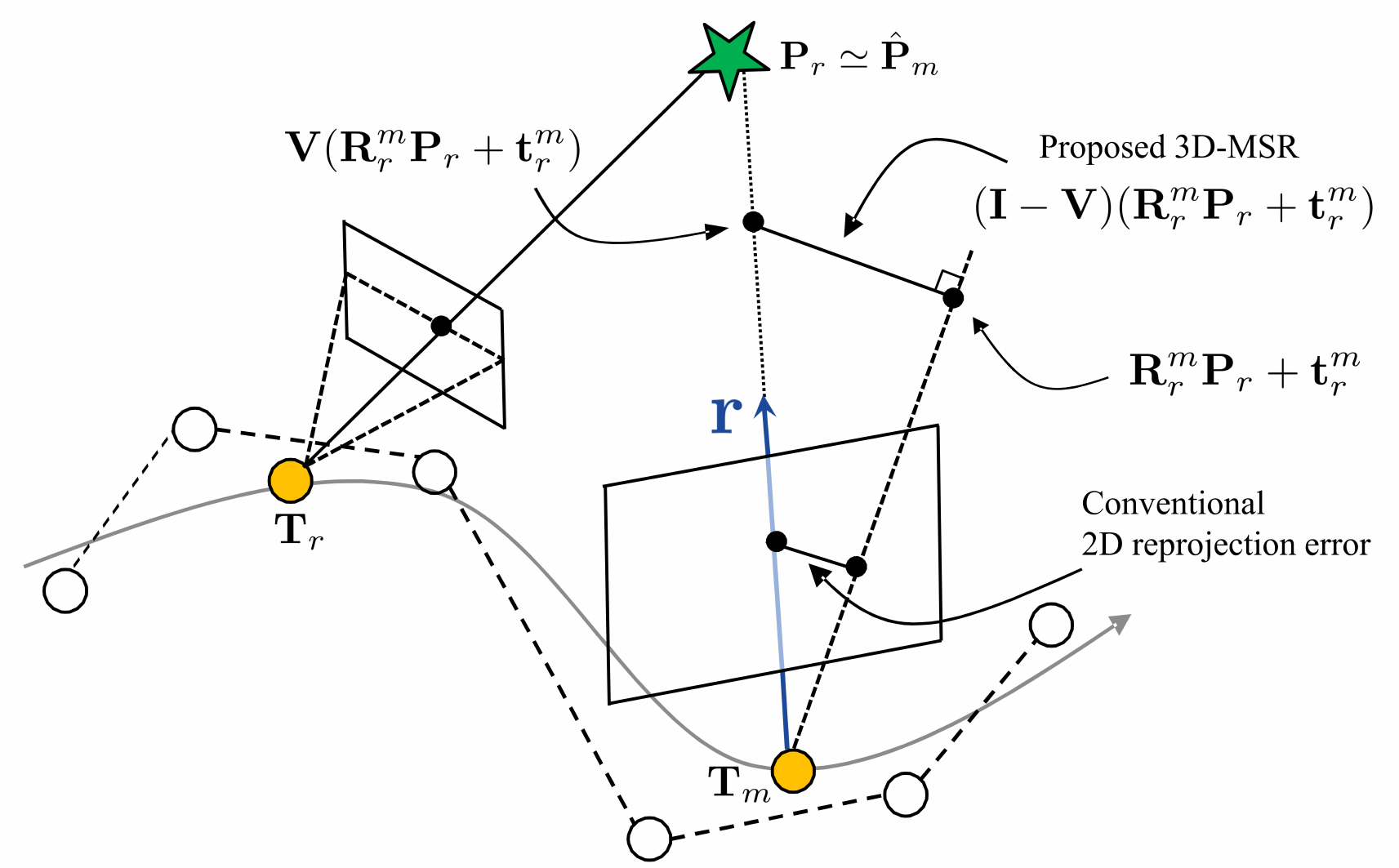}
	\caption{Conceptual illustration of 3D-MSR.
		The reference 3D feature \(\mathbf{P}_r\) is transformed into the measurement frame and compared with the observation ray defined by \(\mathbf{P}_m\).
		White circles indicate AKP knots, yellow circles denote MRS-interpolated poses, and the green star denotes the reference feature.
		The residual imposes a 3D ray-consistency constraint without directly minimizing 2D reprojection error.}
	\label{fig:reference_frame_error}
    \vspace{-0.9cm}
\end{figure}

Let \( \mathbf{P}_r \in \mathbb{R}^3 \) denote the reference 3D feature position, obtained from multi-view triangulation of image correspondences and refined by global bundle adjustment.
Let \( \mathbf{P}_m \in \mathbb{R}^3 \) denote the back-projected measurement-ray direction obtained from the calibrated camera model.
Here, \(\mathbf{P}_m\) defines only the calibrated observation direction and does not contain feature-depth information.
Given the relative transformation \( \mathbf{T}_r^m = \mathbf{T}_m^{-1} \mathbf{T}_r \in SE(3) \), which maps the reference frame to the measurement frame, the residual is formulated as follows:
\vsequ \begin{equation}
	\begin{split}
		\mathbf{f}_P(\mathbf{P}_m, \mathbf{P}_r; \mathbf{T}_r^m, \mathbf{T}_w^s(t)) = (\mathbf{I} - \mathbf{V}) (\mathbf{R}_r^m \mathbf{P}_r + \mathbf{t}_r^m),
	\end{split}
	\label{eqa:3DMSR}
\end{equation} 
\noindent where \( \mathbf{R}_r^m \) and \( \mathbf{t}_r^m \) denote the rotation and translation of \( \mathbf{T}_r^m \), \( \mathbf{I} \in \mathbb{R}^{3 \times 3} \) is the identity matrix, and \( \mathbf{V} \in \mathbb{R}^{3\times3} \) is the projection matrix onto the unit observation-ray direction \( \mathbf{r} \) as follows:
\vsequ \begin{equation}
	\begin{split}
        \mathbf{V} = \frac{\mathbf{r}\mathbf{r}^\top}{\|\mathbf{r}\|^2}, \quad 
        \mathbf{r} = \frac{\mathbf{P}_m}{\|\mathbf{P}_m\|} .
	\end{split}
\end{equation} 
\noindent The normalization defines the unit observation-ray direction \(\mathbf{r}\) and makes the projection independent of the scale of \(\mathbf{P}_m\).
Thus, \(\mathbf{I}-\mathbf{V}\) removes the ray-parallel component and evaluates the ray-orthogonal consistency error in 3D measurement space.
Since the along-ray component is not constrained, 3D-MSR does not make feature depth observable under pure rotation or insufficient parallax.
The depth of \(\mathbf{P}_r\) is instead inherited from prior multi-view triangulation and bundle adjustment, which provides the reference feature when sufficient parallax is available.
The inertial residuals and the smoothness prior in~\eqref{eqa:FULLCOST} regularize the trajectory, while 3D-MSR provides the visual bearing-consistency constraint optimized.
% within~\eqref{eqa:FULLCOST}.
Residuals in~\eqref{eqa:3DMSR} are evaluated on the multi-resolution, non-uniform B-spline trajectory described in \secref{sec:mrs}.
\figref{fig:reference_frame_error} illustrates how the spline-based trajectory provides temporally aligned interpolation for evaluating 3D-MSR.
% This residual is integrated into the loss function in~\eqref{eqa:FULLCOST} and provides the visual constraint used in the optimization.

\subsection{Trajectory Representation Using Multi-Resolution Splines (MRS)}
\label{sec:mrs}
We represent the continuous-time sensor trajectory \(\mathbf{T}_w^s(t)\in SE(3)\) using non-uniform B-spline interpolation over multiple resolution levels.
The spline coefficients are expressed as local pose coordinates and mapped to \(SE(3)\) through Lie group retraction, so the trajectory remains a valid rigid-body motion. 
Let \(\mathcal{H}=\{0,1,\ldots,L\}\) be the set of resolution levels, where \(h=0\) is the coarsest level and \(h>0\) denotes a refinement level.
The MRS trajectory is defined as follows:
\vsequ \begin{equation}
\begin{split}
\mathbf{T}_w^s(t)
=
\mathcal{R}_{SE(3)}\!\left(
\sum_{h\in\mathcal{H}}
\sum_{\mu=0}^{n_h-1}
 w_{\mu}^{h}(t)\mathbf{c}_{\mu}^{h}
\right).
\end{split}
\label{eqa:MRS}
\end{equation}
\noindent Here, \(\mathbf{T}_w^s(t)\) is the pose of the sensor frame \(s\) with respect to the world frame \(w\) at time \(t\), and \(\mathcal{R}_{SE(3)}(\cdot)\) maps the interpolated local coordinate to a valid pose on \(SE(3)\) through the exponential-map update used in the optimizer.
The weighted summation in~\eqref{eqa:MRS} is therefore performed in the local coordinate representation before retraction, not directly on homogeneous transformation matrices.
The index \(h\) denotes the resolution level, \(L\) is the highest refinement level, \(\mu\) indexes a local coefficient, and \(n_h\) is the number of coefficients at level \(h\).
The vector \(\mathbf{c}_{\mu}^{h}\in\mathbb{R}^{6}\) is the local pose coefficient for the \(\mu\)-th support region, and \(w_{\mu}^{h}(t)\) is its effective interpolation weight.

The effective interpolation weight is constructed from fixed B-spline basis functions and bounded refinement coefficients as follows:
\vsequ \begin{equation}
\begin{split}
w_{\mu}^{h}(t)
&= B_{\mu,k}^{h}(t) \\
&\quad + \gamma_{\mu}^{h}
\frac{t-t_{\mu}^{h}}{t_{\mu+1}^{h}-t_{\mu}^{h}}
B_{\mu,k-1}^{h}(t) \\
&\quad + \gamma_{\mu+1}^{h}
\frac{t_{\mu+1}^{h}-t}{t_{\mu+1}^{h}-t_{\mu}^{h}}
B_{\mu+1,k-1}^{h}(t).
\end{split}
\label{eqa:mrs_weight}
\end{equation}
\noindent Here, \(B_{\mu,k}^{h}(t)\), \(B_{\mu,k-1}^{h}(t)\), and \(B_{\mu+1,k-1}^{h}(t)\) are B-spline basis functions defined by the level-\(h\) knot vector and spline order \(k\).
The timestamps \(t_{\mu}^{h}\) and \(t_{\mu+1}^{h}\) are consecutive knots associated with the local support of \(\mathbf{c}_{\mu}^{h}\).
The coefficients \(\gamma_{\mu}^{h}\) and \(\gamma_{\mu+1}^{h}\) activate adjacent refinement terms according to local motion variation.
Accordingly, \(w_{\mu}^{h}(t)\) preserves the fixed spline support while adding bounded local flexibility to the interpolation weight.

The refinement activation is computed from a locally filtered velocity-change statistic as follows:
\vsequ \begin{equation}
\begin{split}
\tilde{\mathbf{v}}_i
&=
\sum_{s=-R}^{R}\kappa_s\mathbf{v}_{i+s}, \\
\delta_{\mu}^{h}
&=
\left\|\tilde{\mathbf{v}}_{i(\mu,h)+1}
-
\tilde{\mathbf{v}}_{i(\mu,h)}\right\|, \\
M_{\Delta\mathbf{v}}^{h}
&=
Q_{0.95}\!\left(\mathcal{W}^{h}\right) .
\end{split}
\label{eqa:mrs_velocity_scale}
\end{equation}
\noindent Here, \(\mathbf{v}_i\) and \(\tilde{\mathbf{v}}_i\) are the raw and locally filtered velocities at the \(i\)-th IMU timestamp, respectively.
The integer \(s\in[-R,R]\) is the filtering offset, \(R\) is the half-window size, and \(\kappa_s\) is a normalized kernel coefficient satisfying \(\sum_{s=-R}^{R}\kappa_s=1\).
The mapping \(i(\mu,h)\) gives the IMU sample index associated with the support of \(\mathbf{c}_{\mu}^{h}\), and \(\delta_{\mu}^{h}\) is the corresponding local velocity-change magnitude.
The set \(\mathcal{W}^{h}\) denotes the filtered velocity-change magnitudes in the current optimization window at level \(h\), and \(Q_{0.95}(\cdot)\) denotes the empirical 95th-percentile operator.

The bounded activation coefficient is defined as follows:
\vsequ \begin{equation}
\begin{split}
\gamma_{\mu}^{h}
=
\operatorname{clip}\!\left(
\frac{\delta_{\mu}^{h}}{M_{\Delta\mathbf{v}}^{h}+\epsilon},
\gamma_{\min}^{h},
\gamma_{\max}^{h}
\right).
\end{split}
\label{eqa:mrs_gamma}
\end{equation}
\noindent Here, \(\operatorname{clip}(\cdot,\gamma_{\min}^{h},\gamma_{\max}^{h})\) bounds its first argument between \(\gamma_{\min}^{h}\) and \(\gamma_{\max}^{h}\), and \(\epsilon>0\) prevents division by zero.
The bounds \(\gamma_{\min}^{h}\) and \(\gamma_{\max}^{h}\) limit abrupt amplification of local refinement terms.
During each linearization, \(\gamma_{\mu}^{h}\) is computed using the propagated velocity estimate and then held fixed, so residual optimization updates the trajectory coefficients under fixed spline support.
Thus, MRS does not introduce a discontinuous change in the active spline basis functions and their associated temporal support during optimization.
Because the refinement term uses lower-order fixed B-spline basis functions, the smoothness of the effective trajectory is bounded by the smoothness of the lowest-order active refinement term under the selected knot multiplicity.
With the spline order in \tabref{tab:parameter_settings}, pose, velocity, and acceleration for inertial residuals are evaluated analytically from the same spline representation.

This hierarchy combines multiple temporal supports within a single continuous-time trajectory representation.
Coarse levels provide long-term smoothness, while finer levels add local detail for sudden turns, vibrations, and nonlinear motion.
Compared with fixed-resolution approaches such as CT-ICP~\cite{dellenbach2022ct} and uniformly spaced continuous-time formulations such as OKVIS-CT~\cite{hug2022continuous}, MRS increases interpolation flexibility in high-variation intervals while maintaining an explicit \(SE(3)\) trajectory representation.
This motion-aware hierarchy supports accurate residual evaluation at arbitrary timestamps without introducing redundant resolution in low-variation intervals.

\subsection{Adaptive Knot Placement (AKP)}
\label{sec:akp}
AKP determines spline knot timestamps from local motion intensity.
High-dynamic intervals receive shorter knot spacings, whereas low-dynamic intervals use longer spacings to avoid unnecessary parameters.
The resulting knot sequence defines the temporal support on which the MRS trajectory in \secref{sec:mrs} is evaluated.

For the \(q\)-th adaptive knot, the timestamp is accumulated from bounded local intervals as follows:
\vsequ \begin{equation}
\begin{split}
t_q &= t_0 + \sum_{r=1}^{q}\Delta t_r, \\
\Delta t_r
&=
\operatorname{clip}\!\left(
\frac{\alpha}{\rho_r+\epsilon},
\Delta t_{\min},
\Delta t_{\max}
\right), \\
\rho_r
&=
\left\|\tilde{\mathbf{v}}_{i_r+1}
-
\tilde{\mathbf{v}}_{i_r}\right\|.
\end{split}
\label{eqa:akp}
\end{equation}
\noindent Here, \(t_q\) is the timestamp of the \(q\)-th adaptive knot, \(t_0\) is the initial knot timestamp, and \(q\) is the adaptive-knot index.
The index \(r\) denotes the local interval accumulated up to knot \(q\), and \(\Delta t_r=t_r-t_{r-1}\) is the duration of the \(r\)-th adaptive interval.
The scalar \(\rho_r\) is the local motion intensity computed from the filtered velocity sequence \(\tilde{\mathbf{v}}_i\) in~\eqref{eqa:mrs_velocity_scale}, and \(i_r\) is the associated IMU sample index.
The gain \(\alpha>0\) maps the motion statistic to knot spacing under a fixed normalization rule, \(\epsilon\) prevents division by zero, and \(\operatorname{clip}(\cdot,\Delta t_{\min},\Delta t_{\max})\) bounds the interval between the minimum and maximum spacings.
The bounds \(\Delta t_{\min}\) and \(\Delta t_{\max}\) are selected from the camera rate, IMU rate, and expected motion bandwidth, not independently tuned for each trajectory.
The same normalization rule is used across sequences to map the motion statistic to the specified knot-interval range.
The inverse relation between \(\rho_r\) and \(\Delta t_r\) increases temporal resolution in high-motion intervals and reduces redundant knots in low-motion intervals.

Unlike uniformly spaced formulations~\cite{furgale2012continuous,hug2022continuous}, AKP allocates temporal resolution according to local motion characteristics.
Compared with CT-UIO's adaptive knot span strategy~\cite{sun2025ct}, AKP uses bounded cumulative intervals to generate a smooth non-uniform knot sequence for the visual-inertial backend.
The resulting trajectory representation provides motion-dependent temporal support for MRS interpolation and 3D-MSR evaluation at visual measurement timestamps.

\section{Experiments}
\label{sec:experiments}
This section evaluates ASPIRE-VINS and quantifies the contributions of AKP, MRS, and 3D-MSR.
The experiments cover highly dynamic motion, realistic visual degradation, and structurally degenerate environments using three complementary data sources, namely a benchmark UAV-style dataset, custom handheld sequences, and selected large-scale architectural sequences.
We report existing VINS comparisons as end-to-end system-level evaluations and use a separate controlled ablation to isolate backend components under identical visual inputs.
Each sequence was evaluated once with fixed parameters and the same recorded stream.
Accordingly, the reported RMSE and maximum-error values are trajectory-level statistics rather than the mean and standard deviation over repeated runs.

We first employ the VIO benchmark dataset~\cite{jeon2021run}, which was originally developed to benchmark VO/VIO algorithms under UAV-like motion patterns.
The dataset provides four canonical trajectory types (\textit{circle}, \textit{infinity}, \textit{square}, and \textit{pure-rotation}), each further divided into \textit{normal}, \textit{fast}, and \textit{head-turning} variants.
These sequences isolate motion-induced failure modes such as high-speed translation and aggressive yaw dynamics.
The \textit{fast} sequences reach translational velocities of approximately 3.0\,m/s, and the \textit{head-turning} cases introduce rapid yaw and parallax variation, making them suitable for evaluating AKP and MRS.

Despite these advantages, the VIO benchmark dataset is inherently limited by its experimental environment.
All sequences are recorded indoors within a compact $3.15 \times 3.60 \times 2.50\,\mathrm{m}^3$ space, under stable illumination and with abundant visual features.
Thus, it does not fully evaluate robustness to visual degradation or structural degeneracy.

To complement these limitations, we further evaluate ASPIRE-VINS on two additional data sources, custom handheld sequences and selected sequences from the Hilti-Oxford dataset~\cite{zhang2022hilti}.
The custom handheld datasets are captured in unconstrained environments and target realistic deployment scenarios that are difficult to reproduce in benchmark settings.
They include abrupt indoor-to-outdoor illumination transitions, feature-deprived regions such as plain white walls and narrow staircases, as well as transparent or reflective surfaces such as glass walls.
These sequences stress robustness to rapid photometric changes and locally weak geometric constraints, but they do not systematically expose the system to long repetitive structures under consistent motion.

To address these aspects, we additionally incorporate selected Hilti-Oxford sequences (\textit{exp04}, \textit{exp05}, \textit{exp06}, and \textit{exp18}).
These sequences contain large-scale layouts with spiral staircases, repetitive structures, and narrow corridors, enabling evaluation under sustained geometric repetition and complementing the localized handheld scenarios.

We collected custom handheld sequences on a system equipped with an Intel\textsuperscript{\textregistered} Core\texttrademark\ i7-7567U processor, 32\,GB of memory, Ubuntu~22.04, and ROS~2 Humble.
The main parameters used in all experiments are summarized in \tabref{tab:parameter_settings}.
These settings are fixed across VIO benchmarks with UAV-like motion, custom handheld data, and Hilti-Oxford data from architectural environments unless otherwise stated.
Knot bounds are determined from the camera rate, IMU rate, and expected motion bandwidth.
The AKP gain maps the robust motion statistic to the specified knot-interval range using the same normalization rule across all evaluations.

\begin{table}[!t]
\vspace{-0.14cm}
\captionsetup{justification=raggedright,singlelinecheck=false}
    \centering
    \caption{Key parameter settings used in the experiments.}
    \label{tab:parameter_settings}
    \vstabcap
    \begingroup
    \footnotesize
    \renewcommand{\arraystretch}{1.08}
    \setlength{\tabcolsep}{3.6pt}
    \resizebox{0.90\columnwidth}{!}{%
    \begin{tabular}{lll}
        \toprule
        \textbf{Parameter} & \textbf{Value} & \textbf{Role} \\
        \midrule
        Camera / IMU rate & 30\,Hz / 400\,Hz & Sensor timing \\
        Spline order $k$ & 4 & $C^2$ trajectory interpolation \\
        Resolution levels $|\mathcal{H}|$ & 3 & Coarse + two refinement levels \\
         Knot bounds $\Delta t_{\min},\Delta t_{\max}$ & $0.03\,\mathrm{s},\,0.25\,\mathrm{s}$ & AKP spacing bounds \\
        AKP gain $\alpha$ & normalized & Motion-to-spacing mapping \\
        $\gamma_{\min}^{h},\gamma_{\max}^{h}$ & 0.0, 1.0 & Bounded MRS activation \\
        Smoothing half-window $R$ & 2 IMU samples & Velocity-change smoothing \\
        Robust upper scale $M_{\Delta\mathbf{v}}^{h}$ & 95th percentile & MRS velocity-change normalization \\
        Stability constant $\epsilon$ & $10^{-6}$ & Division-by-zero prevention \\
        Smoothness weight $\lambda$ & 0.05 & Motion regularization \\
        \bottomrule
    \end{tabular}}
    \endgroup
    % {\scriptsize \parbox{0.901\linewidth}{\vstabfoot Values denote the settings used unless otherwise specified. Knot bounds are determined from the camera rate, IMU rate, and expected motion bandwidth. The AKP gain maps the robust motion statistic to the specified knot-interval range using the same normalization rule across sequences.}}
    \vspace{-0.70cm}
\end{table}

\subsection{Comparison With State-of-the-Art Algorithms}
\label{sec:comp_sota}
We compare ASPIRE-VINS with five representative VINS systems in an end-to-end system-level evaluation, where each method uses its own visual frontend and estimator pipeline.
MSCKF-DVIO~\cite{jung2024msckf} provides a discrete-time filtering reference with Lie group-based IMU propagation, whereas PL-VINS~\cite{fu2020plvins} represents a real-time feature-enhanced VINS that exploits both point and line measurements.
$\sqrt{\mathrm {VINS}}$~\cite{peng2025sqrt} serves as a modern square-root filtering baseline, allowing the runtime and numerical-efficiency characteristics of ASPIRE-VINS to be compared with an optimized discrete-time estimator.
Ctrl-VIO~\cite{lang2022ctrl} and OKVIS-CT~\cite{hug2022continuous} are included as fixed-resolution continuous-time baselines for evaluating the effect of motion-adaptive trajectory representation.
This comparison therefore reports practical system-level performance rather than isolating a single backend component, and the controlled ablation in \secref{sec:controlled_ablation} separately evaluates the effects of the temporal backend and residual formulation.

\subsection{Benchmark Evaluation on VIO Benchmark Dataset}
\label{sec:benchmark_evaluation_on_kaist_vio}
\subsubsection{Pose Estimation Accuracy}

\begin{table*}[!t]
\vspace{-0.14cm}
    \centering
    \caption{Absolute trajectory error (ATE) comparison across nine VIO benchmark sequences.}
    \label{tab:pose_comparison}
    \vstabcap

    \begingroup
    \footnotesize
    \renewcommand{\arraystretch}{0.88}
    \setlength{\tabcolsep}{10pt}

    \setlength{\heavyrulewidth}{0.5pt}
    \setlength{\lightrulewidth}{0.3pt}
    \setlength{\aboverulesep}{0pt}
    \setlength{\belowrulesep}{0pt}

    \resizebox{0.95\textwidth}{!}{%
   \begin{tabular}{l *{12}{N}}
        \toprule
        {Sequence} 
        & \multicolumn{2}{c}{\centering {MSCKF-DVIO}~\cite{jung2024msckf}}
        & \multicolumn{2}{c}{\centering {PL-VINS}~\cite{fu2020plvins}}
        & \multicolumn{2}{c}{\centering {$\sqrt{\mathrm {VINS}}$}~\cite{peng2025sqrt}}
        & \multicolumn{2}{c}{\centering {Ctrl-VIO}~\cite{lang2022ctrl}}
        & \multicolumn{2}{c}{\centering {OKVIS-CT}~\cite{hug2022continuous}}
        & \multicolumn{2}{c}{\centering {ASPIRE-VINS}} \\
        \cmidrule(lr){2-3}\cmidrule(lr){4-5}\cmidrule(lr){6-7}\cmidrule(lr){8-9}\cmidrule(lr){10-11}\cmidrule(lr){12-13}
        & RMSE & Max
        & RMSE & Max
        & RMSE & Max
        & RMSE & Max 
        & RMSE & Max 
        & RMSE & Max \\
        \midrule
        \textit{square-normal}   & 0.083 & 0.217 & 0.093 & 0.245 & 0.086 & 0.231 & \textbf{0.072} & \textbf{0.189} & 0.078 & 0.196 & \underline{0.074} & \underline{0.191} \\
        \textit{square-fast}     & 0.188 & 0.463 & 0.204 & 0.500 & 0.169 & 0.436 & 0.145 & 0.391 & \underline{0.135} & \underline{0.372} & \textbf{0.111} & \textbf{0.345} \\
        \textit{square-head}     & 0.322 & 0.897 & 0.347 & 0.947 & 0.286 & 0.817 & 0.254 & 0.738 & \underline{0.231} & \underline{0.693} & \textbf{0.215} & \textbf{0.688} \\
        \textit{circle-normal}   & 0.093 & 0.254 & 0.100 & 0.277 & 0.089 & 0.241 & 0.085 & 0.222 & \textbf{0.076} & \textbf{0.208} & \underline{0.078} & \underline{0.211} \\
        \textit{circle-fast}     & 0.212 & 0.553 & 0.233 & 0.588 & 0.190 & 0.507 & 0.163 & 0.447 & \underline{0.152} & \underline{0.423} & \textbf{0.137} & \textbf{0.400} \\
        \textit{circle-head}     & 0.388 & 1.124 & 0.414 & 1.181 & 0.347 & 1.040 & 0.309 & 0.965 & \underline{0.290} & \underline{0.920} & \textbf{0.259} & \textbf{0.888} \\
        \textit{infinity-normal} & 0.366 & 1.524 & 0.399 & 1.533 & 0.338 & 1.448 & 0.291 & 1.342 & \textbf{0.269} & \textbf{1.239} & \underline{0.273} & \underline{1.250} \\
        \textit{infinity-fast}   & 0.472 & 1.634 & 0.502 & 1.697 & 0.424 & 1.535 & 0.377 & 1.417 & \textbf{0.339} & \textbf{1.317} & \underline{0.343} & \underline{1.321} \\
        \textit{infinity-head}   & 0.593 & 1.893 & 0.633 & 1.967 & 0.535 & 1.750 & 0.494 & 1.600 & \underline{0.450} & \underline{1.520} & \textbf{0.418} & \textbf{1.479} \\
       \midrule
        Mean & 0.302 & 0.951 & 0.325 & 0.993 & 0.274 & 0.889 & 0.243 & 0.812 & \underline{0.224} & \underline{0.765} & \textbf{0.212} & \textbf{0.753} \\
        \bottomrule
    \end{tabular}}
    \endgroup

    {\scriptsize
      \parbox{0.905\linewidth}{\vstabfoot
      The best performance in each sequence is highlighted in \textbf{bold}, the second best is \underline{underlined}, and all errors are reported in meters [m].}
    }
    \vspace{-0.65cm}
\end{table*}
	
\tabref{tab:pose_comparison} summarizes the absolute trajectory error (ATE) results across nine VIO benchmark sequences, which include \textit{square}, \textit{circle}, and \textit{infinity} trajectories, each executed under \textit{normal}, \textit{fast}, and \textit{head-turning} conditions. 
MSCKF-DVIO performs competitively in structured, low-dynamic motion but degrades under fast or rotational trajectories due to its fixed temporal state representation. 
PL-VINS achieves moderate gains through line constraints, yet its image-plane residuals remain sensitive to feature localization errors under rapid motion.
$\sqrt{\mathrm {VINS}}$ provides an efficient square-root filtering baseline and improves over the conventional discrete-time baselines in most sequences, reflecting its numerical stability and filtering efficiency.
Ctrl-VIO and OKVIS-CT, as fixed-resolution continuous-time baselines, improve temporal consistency but use uniform temporal support, which can be less effective during head-turning or rapidly varying motion.
In contrast, ASPIRE-VINS improves accuracy in high-dynamic and head-turning sequences by combining motion-adaptive spline resolution with 3D measurement-space bearing consistency.

ASPIRE-VINS achieves the lowest mean RMSE of \textbf{0.212\,m} across all sequences, representing a \textbf{5.36\% improvement} over the best baseline, OKVIS-CT.
While Ctrl-VIO slightly surpasses ASPIRE-VINS in the \textit{square-normal} case and OKVIS-CT performs comparably in smooth \textit{infinity} trajectories, adaptive modeling yields clear advantages in nonlinear and rapidly varying motion. 
These results indicate that ASPIRE-VINS is most effective when temporal resolution and residual evaluation must adapt to motion complexity.

\subsubsection{Controlled Backend Ablation}
\label{sec:controlled_ablation}
To reduce frontend confounding, all variants in \tabref{tab:controlled_backend_ablation} are implemented within the same feature-processing pipeline and use identical feature tracks, triangulated landmarks, and outlier rejection.
Only the trajectory representation, knot strategy, and residual formulation differ across the ablation variants; therefore, the controlled-backend values are reported as a separate analysis and are not expected to numerically match the end-to-end system-level results in \tabref{tab:pose_comparison}.
The selected sequences include low-dynamic motion (\textit{square-normal}), translation-heavy motion (\textit{square-fast}), and pure-rotation cases (\textit{rotation-normal} and \textit{rotation-fast}), which are used to assess ray-orthogonal consistency rather than depth observability.

\begin{table}[!t]
    \vspace{-0.14cm}
    \centering
    \caption{Controlled ablation using identical visual feature tracks.}
    \label{tab:controlled_backend_ablation}
    \vstabcap
    \begingroup
    \footnotesize
    \renewcommand{\arraystretch}{1.0}
    \setlength{\tabcolsep}{3.5pt}

    \setlength{\heavyrulewidth}{0.6pt}
    \setlength{\lightrulewidth}{0.4pt}
    \setlength{\aboverulesep}{0pt}
    \setlength{\belowrulesep}{0pt}

    \resizebox{0.95\columnwidth}{!}{%
    \begin{tabular}{lccccccc}
        \toprule
        {Variant} & {CT} & {A/M} & {MSR}
        & \textit{sq.-n.} & \textit{sq.-f.}
        & \textit{rot.-n.} & \textit{rot.-f.} \\
        \midrule
        DT + 2D reproj.       & \xmark & \xmark & \xmark      & 0.092 & 0.204 & 0.814 & 0.872 \\
        DT + 3D-MSR           & \xmark & \xmark & \checkmark  & 0.087 & 0.179 & 0.746 & 0.803 \\
        CT uni. + 2D reproj.  & \checkmark & \xmark & \xmark      & 0.090 & 0.187 & 0.823 & 0.889 \\
        CT uni. + 3D-MSR      & \checkmark & \xmark & \checkmark  & 0.083 & 0.156 & 0.681 & 0.736 \\
        CT A/M + 2D reproj.   & \checkmark & \checkmark & \xmark  & 0.085 & 0.166 & 0.728 & 0.782 \\
        CT-UIO span + 3D-MSR~\cite{sun2025ct}
                               & \checkmark & $\sim$ & \checkmark  & 0.080 & 0.143 & 0.639 & 0.699 \\
        Full proposed backend           & \checkmark & \checkmark & \checkmark
                               & \textbf{0.076} & \textbf{0.126} & \textbf{0.552} & \textbf{0.612} \\
        \bottomrule
    \end{tabular}}
    \endgroup
    {\scriptsize \parbox{0.92\columnwidth}{\vstabfoot ATE RMSE is reported in meters [m] for the internal controlled-backend ablation.
    ``DT,'' ``CT,'' ``uni.,'' ``A/M,'' ``MSR,'' and ``reproj.'' denote a preintegrated discrete-time backend,
    the continuous-time backend, uniform knot spacing, AKP/MRS, 3D measurement-space residuals, and 2D reprojection residuals, respectively.
    ``sq.-n.,'' ``sq.-f.,'' ``rot.-n.,'' and ``rot.-f.'' denote the \textit{square-normal},
    \textit{square-fast}, \textit{rot-normal}, and \textit{rot-fast} sequences, respectively.
    The symbol $\sim$ denotes replacement of AKP with the CT-UIO-style knot-span rule.
    The best values are highlighted in \textbf{bold}.}}
    \vspace{-0.80cm}
\end{table}

Because all rows share the same feature tracks, the DT rows separate the effect of 3D-MSR from the continuous-time representation, while the CT rows separate temporal modeling from the residual formulation.
Adding 3D-MSR improves both DT and CT variants by replacing pixel-plane reprojection with a bearing-consistency constraint evaluated in 3D measurement space.
The rotation-dominant results reflect improved ray-orthogonal consistency and reduced sensitivity to image-plane perturbations under calibrated observations, rather than a separate depth-observability or calibration-error compensation effect.
Comparing CT uniform + 3D-MSR with the full model shows that AKP/MRS mainly improves fast translation and rotation-heavy motion, while the small change in \textit{square-normal} indicates that the adaptive representation remains stable in low-dynamic motion.
The CT-UIO-style variant changes knot spans only, whereas ASPIRE-VINS couples bounded AKP intervals with MRS local refinement.
Its lower errors support this combined refinement within the visual-inertial backend.

\subsubsection{Runtime Analysis}
\tabref{tab:runtime_comparison} summarizes the average processing time over the nine VIO benchmark sequences using the same input data and CPU platform.
Among the evaluated methods, $\sqrt{\mathrm {VINS}}$ achieves the lowest latency of 18\,ms per frame ($\approx$56\,FPS), reflecting the computational efficiency of a square-root filter-based discrete-time estimator.
ASPIRE-VINS requires 56\,ms per frame ($\approx$18\,FPS), which is slower than the optimized filtering baselines but faster than OKVIS-CT and comparable to Ctrl-VIO.
This runtime indicates that the proposed backend maintains practical processing speed while providing motion-adaptive continuous-time modeling.

\begin{table}[!t]
    \vspace{-0.14cm}
    \captionsetup{justification=raggedright, singlelinecheck=false}
	\centering
	\caption{Runtime comparison of VINS algorithms averaged across all nine VIO benchmark sequences under the same evaluation platform.}
	\label{tab:runtime_comparison}
    \vstabcap
    \begingroup
    \footnotesize
    \renewcommand{\arraystretch}{0.92}
    \setlength{\tabcolsep}{15pt}
    \setlength{\heavyrulewidth}{0.6pt}
    \setlength{\lightrulewidth}{0.4pt}
    
    \setlength{\aboverulesep}{0pt}
    \setlength{\belowrulesep}{0pt}

    \resizebox{0.95\columnwidth}{!}{
    \begin{tabular}{lccc}
    \toprule
    Method & 
    \makecell{Avg. time [ms]} & 
    FPS & 
    \makecell{\scriptsize Adaptive\\[-1pt]\scriptsize resolution} \\
    \midrule
    MSCKF-DVIO~\cite{jung2024msckf} & 38 & $\approx$26 & \xmark  \\
    PL-VINS~\cite{fu2020plvins} & 44 & $\approx$23 & \xmark  \\
    $\sqrt{\mathrm {VINS}}$~\cite{peng2025sqrt} & \textbf{18} & \textbf{$\approx$56} & \xmark  \\
    \rowcolor{gray!20} Ctrl-VIO~\cite{lang2022ctrl} & 63 & $\approx$16 & \xmark  \\
    \rowcolor{gray!20} OKVIS-CT~\cite{hug2022continuous} & 87 & $\approx$11 & \xmark  \\
    \rowcolor{gray!20} ASPIRE-VINS & 56 & $\approx$18 & \checkmark  \\
    \bottomrule
\end{tabular}}
    \endgroup
    {\scriptsize \parbox{0.92\linewidth}{\vstabfoot Continuous-time methods are shaded, and the best values are highlighted in \textbf{bold}. 
    All values were measured on the same Intel Core i7-7567U CPU platform with 32\,GB memory under Ubuntu 22.04/ROS~2 Humble, using the same input resolution and nine-sequence set.
    The $\sqrt{\mathrm {VINS}}$ result is obtained from its released ROS~2 implementation and includes the visual-inertial estimator pipeline used in this comparison.}}
    \vspace{-0.75cm}
\end{table}

\subsection{Evaluation on Custom Handheld Datasets}
\subsubsection{Sensor Configuration}
\begin{figure}[!ht]
	  \vspace{-0.3cm}
    \centering
	\includegraphics[width=0.17\textwidth]{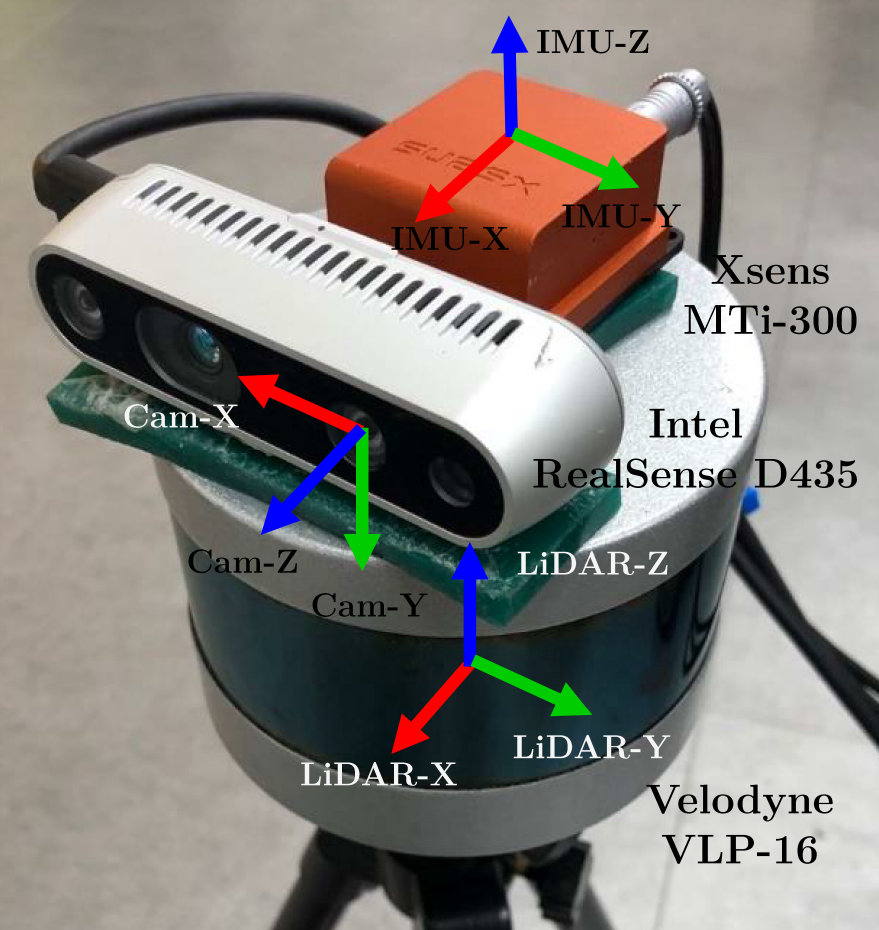}
	\caption{Hardware configuration for custom handheld data collection.}
	\label{fig:sensor_configuration}
    \vspace{-0.5cm}
\end{figure}

As discussed in \secref{sec:experiments}, we collected custom handheld sequences on our campus using the multi-sensor platform shown in \figref{fig:sensor_configuration}. 

The platform consists of an Intel RealSense D435 RGB-D camera, an Xsens MTi-300 IMU, and a Velodyne VLP-16 LiDAR with rigid mounting and hardware-level synchronization.
All custom datasets were recorded at 30\,Hz for the camera, 400\,Hz for the IMU, and 10\,Hz for the LiDAR.
Accordingly, the custom sequences are used to evaluate trajectory accuracy and robustness under realistic visual and geometric degradation.

\subsubsection{Ground Truth via PALoc}
To generate accurate 6-DoF ground truth, we employ PALoc~\cite{hu2024paloc}, a prior-assisted localization framework that fuses LiDAR odometry, IMU, and prior map constraints in a factor-graph optimization.
Unlike conventional scan-to-map pipelines~\cite{cwian2021large,zhou2021s4}, PALoc provides frame-wise pose and uncertainty estimates, which are useful in degenerate indoor geometries such as corridors and staircases.
PALoc poses are aligned and interpolated to the VIO timestamps, and global consistency is checked using map continuity and uncertainty trends around transitions such as stairs and doorways.
We also report maximum trajectory errors alongside RMSE to reflect worst-case deviations at elevation changes and abrupt illumination shifts.

\begin{table*}[!t]
    \vspace{-0.14cm}
    \centering
    \caption{Absolute trajectory error (ATE) comparison on custom handheld and public Hilti-Oxford datasets.}
    \label{tab:unified_pose_comparison}
    \vstabcap

    \begingroup
    \footnotesize
    \renewcommand{\arraystretch}{0.88}
    \setlength{\tabcolsep}{10pt}

    \setlength{\heavyrulewidth}{0.5pt}
    \setlength{\lightrulewidth}{0.3pt}
    \setlength{\aboverulesep}{0pt}
    \setlength{\belowrulesep}{0pt}

    \resizebox{0.95\textwidth}{!}{%
    \begin{tabular}{l *{12}{N}}
        \toprule
        {Sequence} 
        & \multicolumn{2}{c}{\centering {MSCKF-DVIO}~\cite{jung2024msckf}}
        & \multicolumn{2}{c}{\centering {PL-VINS}~\cite{fu2020plvins}}
        & \multicolumn{2}{c}{\centering {$\sqrt{\mathrm {VINS}}$}~\cite{peng2025sqrt}}
        & \multicolumn{2}{c}{\centering {Ctrl-VIO}~\cite{lang2022ctrl}}
        & \multicolumn{2}{c}{\centering {OKVIS-CT}~\cite{hug2022continuous}}
        & \multicolumn{2}{c}{\centering {ASPIRE-VINS}} \\
        \cmidrule(lr){2-3}\cmidrule(lr){4-5}\cmidrule(lr){6-7}\cmidrule(lr){8-9}\cmidrule(lr){10-11}\cmidrule(lr){12-13}
        & RMSE & Max
        & RMSE & Max
        & RMSE & Max
        & RMSE & Max 
        & RMSE & Max 
        & RMSE & Max \\
        \midrule
        \textit{Multistory Stairs}   & 0.680 & 0.954 & 0.480 & 0.720 & 0.408 & 0.846 & \underline{0.320} & \underline{0.500} & 0.340 & 0.540 & \textbf{0.240} & \textbf{0.420} \\
        \hline
        \textit{Indoor-to-Outdoor}     & 0.707 & 1.004 & 0.491 & 0.754 & 0.418 & 0.774 & 0.333 & 0.524 & \textbf{0.308} & \underline{0.500} & \underline{0.312} & \textbf{0.448} \\
        \hline
        \textit{HILTI-OXFORD exp04}     & 1.709 & 3.175 & 1.481 & 2.854 & 1.234 & 2.490 & 1.025 & \underline{1.589} & \underline{0.854} & 1.635 & \textbf{0.570} & \textbf{0.964} \\
        \textit{HILTI-OXFORD exp05}   & 1.571 & 2.995 & 1.362 & 2.047 & 1.136 & 2.115 & 0.943 & \underline{1.728} & \underline{0.786} & 1.894 & \textbf{0.524} & \textbf{0.784} \\
        \textit{HILTI-OXFORD exp06}     & 2.091 & 4.057 & 1.812 & 3.229 & 1.512 & 2.865 & 1.255 & 1.954 & \underline{1.046} & \underline{1.671} & \textbf{0.697} & \textbf{1.136} \\
        \textit{HILTI-OXFORD exp18}   & 1.136 & 2.544 & 0.984 & 1.764 & 0.805 & 1.435 & 0.681 & 1.065 & \underline{0.568} & \underline{1.020} & \textbf{0.379} & \textbf{0.690} \\
        \hline\hline
        Mean & 1.316 & 2.455 
        & 1.102 & 1.895 
        & 0.919 & 1.754 
        & 0.760 & 1.227  
        & \underline{0.650} & \underline{1.210} 
        & \textbf{0.454} & \textbf{0.740} \\
        \bottomrule
    \end{tabular}}
    \endgroup

    {\scriptsize
      \parbox{0.905\linewidth}{\vstabfoot
      The best performance in each sequence is highlighted in \textbf{bold}, the second best is \underline{underlined}, and all errors are reported in meters [m].}
    }
    \vspace{-0.4cm}
\end{table*}

\begin{figure*}[!th]
    \vspace{-0.14cm}
    \centering
    \footnotesize
    \begin{minipage}{0.39\textwidth}
        \centering
        \includegraphics[width=\linewidth]{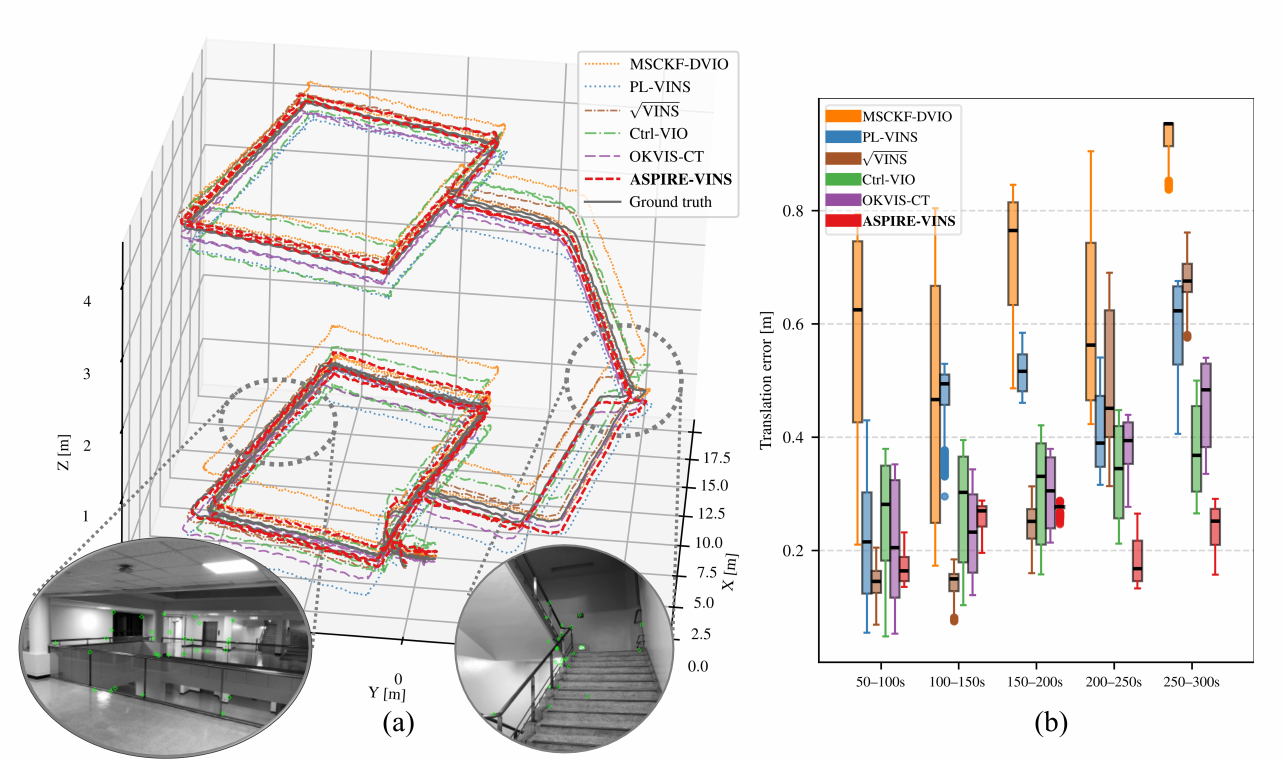}
    \end{minipage}
    \hspace{0.08\textwidth}
    \begin{minipage}{0.39\textwidth}
        \centering
        \includegraphics[width=\linewidth]{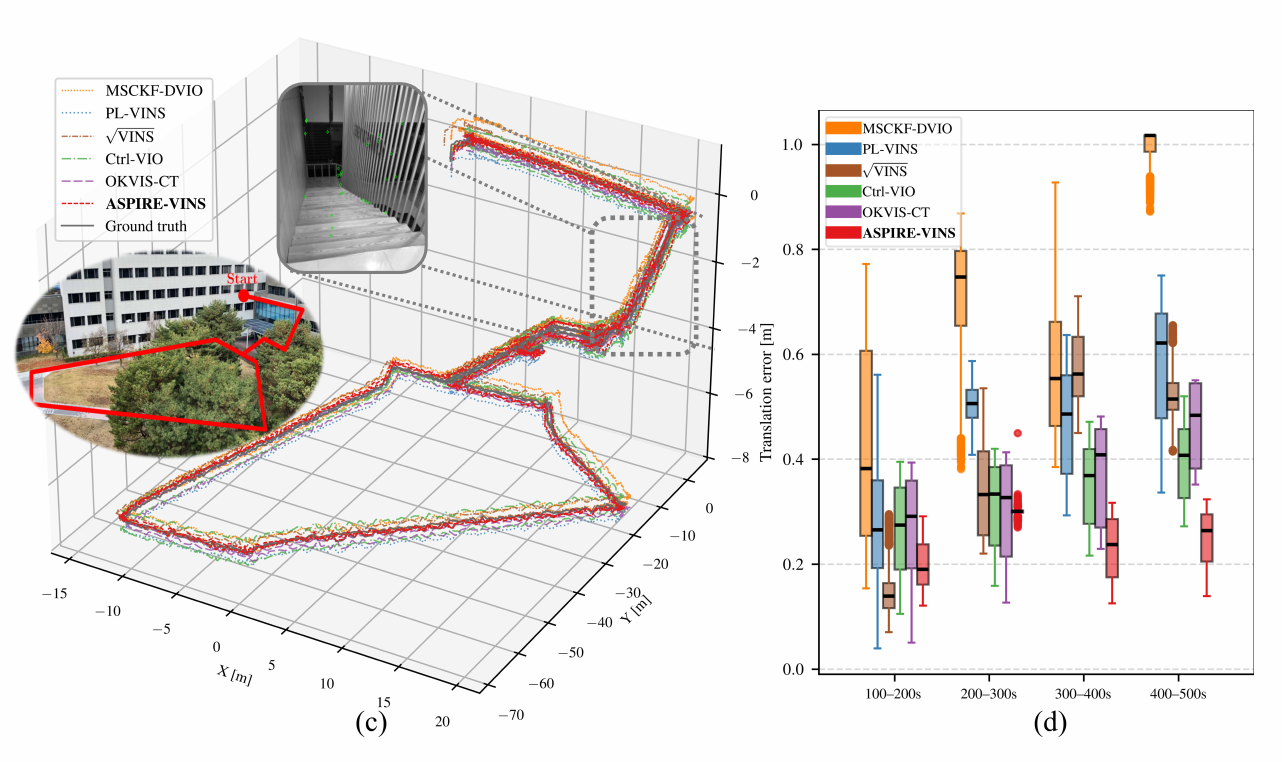}
    \end{minipage}
    \caption{Evaluation on custom handheld datasets with PALoc ground truth.
    (\textbf{a}) \textit{Multistory Stairs} sequence showing the full 3D trajectory with representative environment snapshots of narrow staircases and feature-sparse regions.
    (\textbf{c}) \textit{Indoor-to-Outdoor Transition} sequence showing the 3D trajectory under abrupt illumination changes, transparent surfaces, and outdoor exposure.
    (\textbf{b}) and (\textbf{d}) show translation error distributions over temporal segments for the two sequences, respectively.}
    \label{fig:kaist_custom_handheld}
    \vspace{-0.8cm}
\end{figure*}

\subsubsection{Pose Estimation Accuracy on Custom Handheld Sequences}
The custom handheld evaluation focuses on visual robustness in realistic deployment conditions, including repetitive stair geometries with limited texture (\textit{Multistory Stairs}) and abrupt photometric transitions across glass boundaries (\textit{Indoor-to-Outdoor}).
As shown in \tabref{tab:unified_pose_comparison}, the discrete-time baselines provide references for efficient filtering and image-plane residual formulations under realistic visual degradation.
MSCKF-DVIO and PL-VINS exhibit larger errors in these sequences, where motion blur, exposure shifts, and feature dropouts reduce the reliability of image-domain constraints.
$\sqrt{\mathrm {VINS}}$ reduces average drift through square-root covariance filtering but remains less accurate than the continuous-time variants in these challenging handheld sequences.

Continuous-time baselines maintain smoother trajectories, but their fixed-resolution splines can be suboptimal when motion and visual quality change locally.
In \textit{Multistory Stairs}, ASPIRE-VINS achieves the lowest RMSE and maximum error.
In \textit{Indoor-to-Outdoor}, OKVIS-CT achieves the lowest RMSE, while ASPIRE-VINS achieves the lowest maximum error.
These results suggest that adaptive temporal resolution is beneficial in locally degraded segments, particularly near stair landings and photometric transition boundaries (\tabref{tab:unified_pose_comparison}).
The supplementary videos are provided only as explanatory visualizations of knot density, MRS activation, and the 3D-MSR residual surrogate, not as independent proof of smoothness, differentiability, or calibration robustness.
This behavior is consistent with \figref{fig:kaist_custom_handheld}, where AKP increases temporal resolution near landings and transition boundaries, MRS preserves global smoothness across the sequence, and 3D-MSR provides bearing consistency under feature-deprived or photometrically unstable conditions.

\subsection{Pose Estimation Accuracy in Hilti-Oxford Datasets}
The custom handheld sequences capture abrupt illumination changes and locally feature-deprived regions, but not prolonged structural repetition.
We therefore evaluate selected Hilti-Oxford sequences (\textit{exp04}, \textit{exp05}, \textit{exp06}, and \textit{exp18}), which contain spiral staircases, repetitive geometry, and narrow corridors.

\begin{figure}[!ht]
\vspace{-0.14cm}
	\centering
	\includegraphics[width=0.29\textwidth]{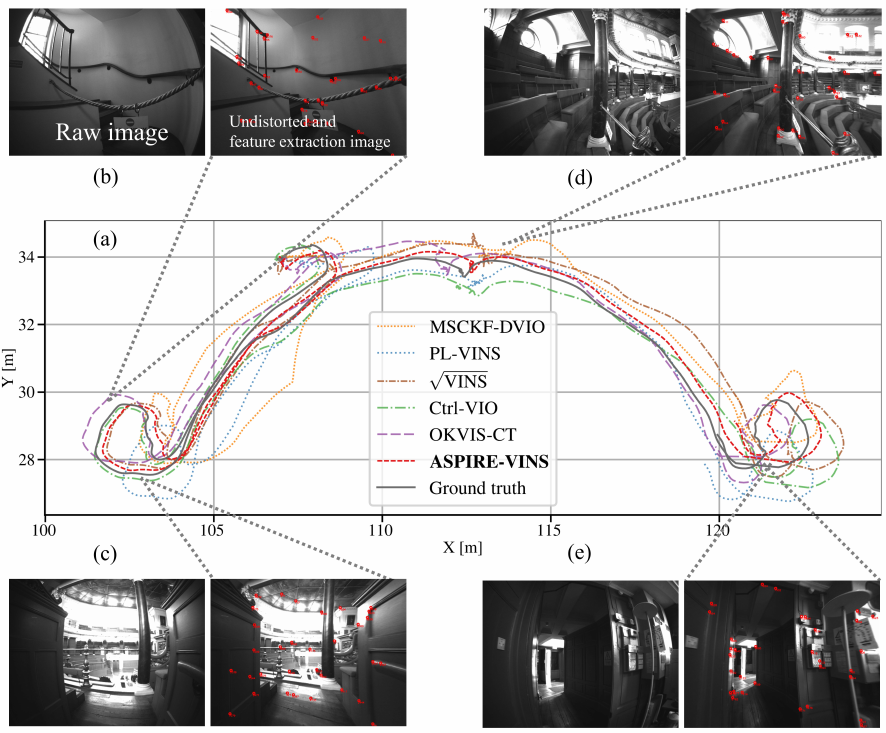}
	\caption{Evaluation on the \textit{Hilti-Oxford exp18} sequence.
	(a) shows the estimated trajectories projected onto the XY plane, highlighting long-term drift behavior under repetitive motion.
	(b) illustrates a narrow spiral staircase with highly repetitive geometry, which induces strong structural degeneracy.
	(c) and (d) depict regions with strong illumination contrast caused by natural lighting, leading to severe photometric degradation and unstable feature observations.
	(e) highlights dark corner regions with limited texture and visibility.}
	\label{fig:hilti_exp18}
    \vspace{-0.9cm}
\end{figure}

As summarized in \tabref{tab:unified_pose_comparison}, ASPIRE-VINS achieves the lowest RMSE and maximum error across all selected sequences.
The advantage is particularly evident in \textit{exp18}, where highly repetitive spiral staircases induce strong structural degeneracy and natural lighting produces severe illumination contrast and dark corner regions (\figref{fig:hilti_exp18}).
This evaluation tests whether a motion-adaptive continuous-time trajectory can preserve consistency through sustained geometric and photometric degradation.
Under these challenges, AKP and MRS provide motion-dependent temporal support, while 3D-MSR contributes bearing consistency under unstable feature observations.

\section{Conclusion}
This letter presented ASPIRE-VINS, a continuous-time visual-inertial odometry framework that combines AKP, MRS, and 3D-MSR within a unified Lie group optimization.
AKP provides motion-dependent knot spacing, MRS adds bounded multi-resolution refinement, and 3D-MSR contributes bearing consistency through calibrated observation-ray constraints in 3D measurement space.
Experiments showed competitive or lower errors than the evaluated baselines.
The controlled ablation showed that temporal adaptivity and 3D-MSR improve accuracy under identical visual inputs, while the runtime analysis confirmed practical speed among continuous-time methods.
The framework remains sensitive to knot initialization and numerical conditioning over long trajectories, and depth observability remains limited under pure rotation or insufficient parallax.
Future work will incorporate Chebyshev-based barycentric rational interpolation (CBRI) for improved stability.
\bibliographystyle{URL-IEEEtrans}
\bibliography{URL-bib}

\end{document}